\documentclass[10pt,twocolumn,a4paper]{article} 
\usepackage{graphicx}
\usepackage{amssymb}
\usepackage{url,hyperref}
\usepackage{array}
\usepackage{grffile}
\usepackage[T1]{fontenc}
\usepackage{abstract}

\usepackage[utf8]{inputenc}
\usepackage{subfigure}
\usepackage{booktabs} 
\usepackage{cellspace, tabularx}
\addparagraphcolumntypes{X}

\usepackage{fancyhdr}
\usepackage[compact]{titlesec}
\titlespacing{\section}{1pt}{*0}{*0}
\usepackage[utf8]{inputenc}

\begin{document}
\setlength{\cellspacetoplimit}{12pt}
\setlength{\cellspacebottomlimit}{12pt}
\title{\vspace{0cm} \bf \Large \rule{\textwidth}{2pt} 
\begin{center}
    \begin{tabularx}{\linewidth}{S{>{\centering\arraybackslash}X}}
        Diagnosis of Acute Myeloid Leukaemia Using Machine Learning \\
    \end{tabularx}
\end{center}
\rule{\textwidth}{2pt}
}

\author{
\begin{tabular}[t]{c@{\extracolsep{8em}}c} 
Athanasios Angelakis  & Ioanna Soulioti \\
\\
JADS & Department of Biology \\ 
Eindhoven University of Technology & University of Athens \\
Den Bosch, Netherlands & Athens, Greece \\
a.angelakis@tue.nl & ioannasoul@biol.uoa.gr
\end{tabular}
\date{}
}

\maketitle    
 
\thispagestyle{empty}

\begin{abstract}
We train a machine learning model on a dataset of 2177 individuals using as features 26 probe sets and their age in order to classify if someone has acute myeloid leukaemia or is healthy. The dataset is multicentric and consists of data from 27 organisations, 25 cities, 15 countries and 4 continents. The accuracy or our model is 99.94\% and its F1-score is 0.9996. To the best of our knowledge the performance of our model is the best one in the literature, as regards the prediction of AML using similar or not data. Moreover, there has not been any bibliographic reference associated with acute myeloid leukaemia for the 26 probe sets we used as features in our model. 
\end{abstract}

\section{Introduction}
Acute myeloid leukaemia (AML) \cite{Short18} is often characterized by non detectable early symptoms and its quick prognosis, even in an intensive care unit could have a huge impact on the overall survival \cite{Mottal20}. The use of machine learning can be helpful on the diagnosis of this disease and therefore in the creation of a screening tool \cite{Roushangar19},  \cite{Abelson18}. Here we focus on the primary diagnosis of AML using the minimum number of probe sets possible in order to achieve excellent performance. In addition, we use the age as feature to our final model since its prognostic value is high regarding the survival of patients with AML \cite{Mosquera21}. Another reason we include the age is that from deep learning work in radiology, in particular in ultrasound with even small data sets of 100 data instances \cite{Angelakis18a}, \cite{Angelakis18b}, and with CatBoost \cite{Angelakis21} using features coming from different sources we can achieve high performance in binary classification problems both on sensitivity and specificity.

We first tune a CatBoost \cite{Prokhorenkova18} on a curated publicly available Affymetrix microarray gene expression and normalized batch corrected dataset consisted of probe sets of 3374 individuals \cite{Roushangar19}, in order to classify if an individual has AML or is healthy. CatBoost library offers the option to return the set of features' importance of CatBoost algorithm and also the set of features' importance of the loss function change. The above two sets can differ. 

We keep the 100 most important features for each of the above two sets and then we take the intersection of these which
consists of 34 probe sets. The idea of intersection comes from the fact that we would like to include features of high importance regarding the predictability of CatBoost algorithm and at the same time its loss function change during the training process.

\begin{figure}[t]
\vskip 0.2in
\begin{center}
\centerline{\includegraphics[width=\columnwidth]{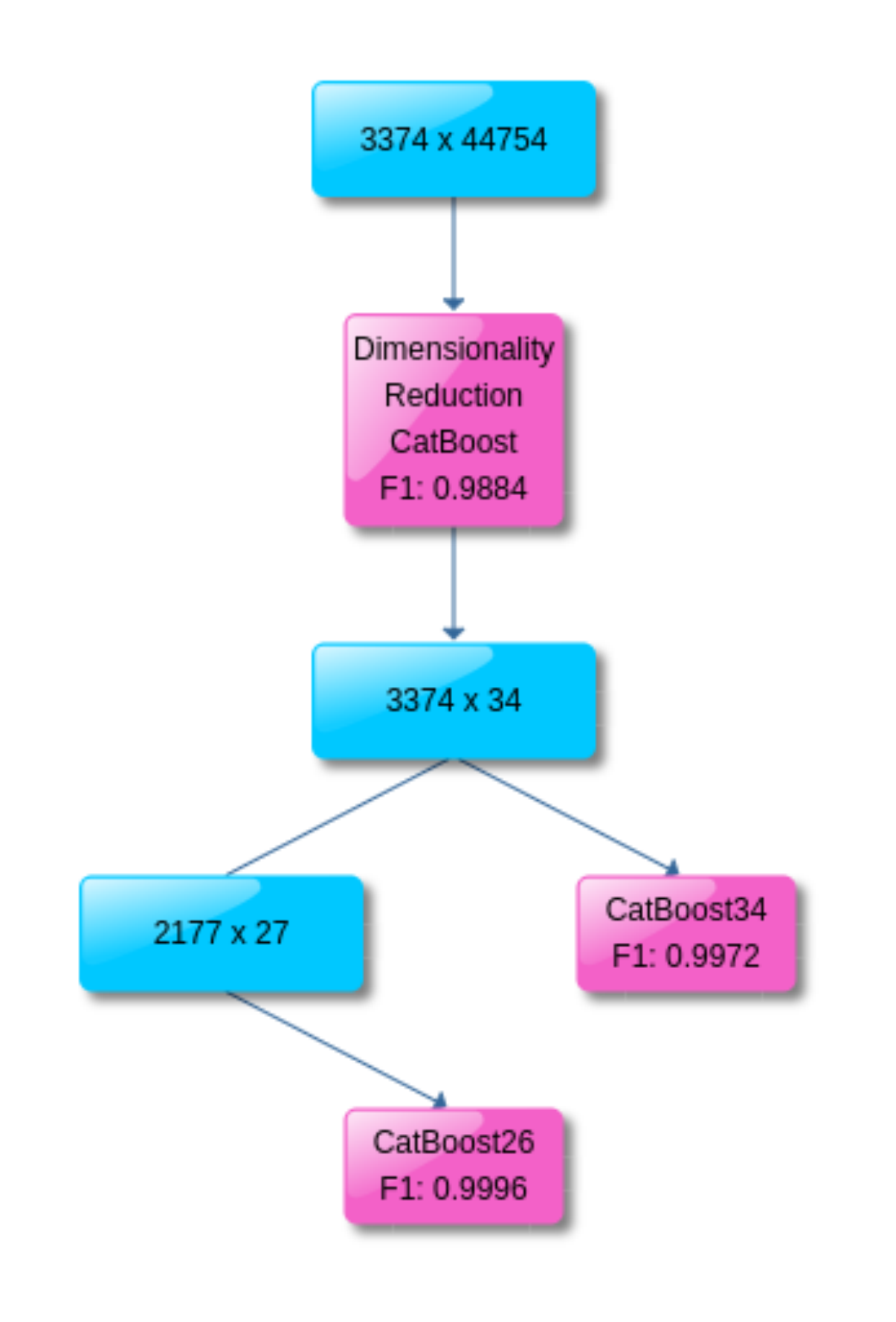}}
\caption{Datasets and CatBoost models with their harmonic mean of precision and recall. The first integer corresponds to the number of the data instances and the second one corresponds to the number of features.}
\label{flow}
\end{center}
\vskip -0.2in
\end{figure}

We randomly split the dataset of the 34 probe sets and the 3374 data instances using 80\% for training and 20\% for validation. We use 10 fold cross validation (10CV) \cite{Kohavi95} in order to tune a CatBoost on the training set, and then we validate it on the test set. 

From these 34 probe sets we keep only those for which we cannot find any bibliographic reference regarding their correlation to AML, Table \ref{t26-probesets}. The only correlated to AML feature we include in our final machine learning model is the age of each individual.

We randomly split the dataset of 2177 individuals using 80\% for training and 20\% for validation. We use 10CV in order to tune the CatBoost on the training set, and then we validate it on the test set. 

In Figure \ref{flow} we show diagram of the three models and the corresponding datasets of our approach.

\section{Models}
The dimensionality reduction CatBoost model has 200 iterators, depth 6 and learning rate 0.1. We randomly split the initial dataset of 3374 data instances and 44754 probe sets. The performance of the tuned model appears in Table \ref{catboost_1-table}.

\begin{table}[t]
\caption{Performance of the dimensionality reduction CatBoost model of the 10CV on the 80\% training set and on the 20\% validation set of 3374 data instances and 44754 probe sets. The dataset corresponds to U133A, U133B and U133 2.0 microarrays.}
\label{catboost_1-table}
\vskip 0.15in
\begin{center}
\begin{small}
\begin{sc}
\begin{tabular}{lcr}
\toprule
Metrics & Validation Set & 10CV \\
\midrule
Spec. & 0.9929 & 0.9805 \\
Sens. & 1.0000 & 0.9991\\
AUC   & 0.9965 & 0.9898\\
F1-score & 0.9964 & 0.9884\\
\bottomrule
\end{tabular}
\end{sc}
\end{small}
\end{center}
\vskip -0.1in
\end{table}

\begin{table}[t]
\caption{Performance of the CatBoost34 model of the 10CV on the 80\% training set and on the 20\% validation set of 3374 data instances and 34 probe sets. The dataset corresponds to U133A, U133B and U133 2.0 microarrays.}
\label{catboost_2-table}
\vskip 0.15in
\begin{center}
\begin{small}
\begin{sc}
\begin{tabular}{lcr}
\toprule
Metrics & Validation Set & 10CV \\
\midrule
Spec. & 1.0000 & 0.9929 \\
Sens. & 1.0000 & 0.9926\\
AUC   & 1.0000 & 0.9920\\
F1-score & 1.0000 & 0.9972\\
\bottomrule
\end{tabular}
\end{sc}
\end{small}
\end{center}
\vskip -0.1in
\end{table}

We compute the intersection of the sets of the most important features, regarding the predictability of CatBoost, and the most important features regarding the loss function change during the training process. We set the number of elements of each set to be 100. The intersection has only 34 probe sets. We tune a CatBoost model (CatBoost34) of 200 iterators, depth 5 and learning rate 0.1 on the dataset of 3374 data instances. The results in Table \ref{catboost_2-table} show that using only 34 probe sets our machine learning model is able to achieve great performance.

From the 34 probe sets we exclude all which are correlated from bibliographic references to AML so we keep only the 26 probe sets of Table \ref{t26-probesets}. The tuned CatBoost model which we use for the diagnosis of AML (CatBoost26) has 100 iterators and depth 11 with learning rate 0.1. 

In all three models above we use the weight balance parameters of CatBoost library since our datasets are imbalanced. Moreover, we keep all the other parameters of them similar to the default values provided by CatBoost library.

\section{Data}
The initial dataset is a curated publicly available Affymetrix microarray gene expression one and it consists of 34 datasets derived from 32 studies \cite{Roushangar19}. It is an international multicentric dataset since its data instances come from 27 organisations, 25 cities, 15 countries and 4 continents. The data come from different transcriptomic platforms: Affymetrix Human Genome U133 Plus 2.0 microarray, Affymetrix Human Genome U133A microarray and Affymetrix Human Genome U133B microarray.

At first, the dataset consisted of 44754 probe sets and 3374 data instances which corresponded to 3374 individuals. From the 3374 data instances 2668 (79.08\%) were labelled as AML and 706 (20.92\%) as healthy. 

The dimensionality reduction tuned model is applied on this dataset. We keep the 26 probe sets of the 34 \{227923\_at, 212549\_at, 219386\_s\_at, 207754\_at, 208022\_s\_at, 209543\_s\_at, 210244\_at, 207206\_s\_at, 210789\_x\_at, 239766\_at, 241688\_at, 244719\_at, 236952\_at, 241611\_s\_at, 217901\_at, 229963\_at, 230527\_at, 222312\_s\_at, 214705\_at, 203294\_s\_at, 209603\_at, 243659\_at, 230753\_at, 204777\_s\_at, 234632\_x\_at, 217680\_x\_at, 219513\_s\_at, 214719\_at, 211772\_x\_at, 207636\_at, 243272\_at, 214945\_at, 226311\_at, 242056\_at\} for which, to the best of our knowledge, there has not been any reference regarding their correlation to AML yet. Since we want to use also the age of the individuals as feature to our diagnostic CatBoost model, we drop-out all the data instances with no age filled-in. 

The final dataset consists of 2177 data instances and it has 27 features (26 probe sets and the age). Tables \ref{trainset2177-table}, \ref{tage} and \ref{info2177-table} provide detailed information about the dataset, including the number of samples used, the sample source, the sex and the age of the individuals, the organisations which provided the data, the AML subtypes and statistics about the overall survival when available, as well as the total number of AML patients and healthy individuals. 

From the 2177 individuals, 1013 are female (46.53\%), 943 are male (43.32\%) and 221 are unknown (10.15\%). In addition, 1629 are AML patients (74.83\%) and 548 are healthy (25.17\%). The mean and the standard deviation of age are 48.87 and 17.01, respectively. As regards the number of data instances per age group in the data set we have: 99 [0-19], 217 [20-29], 340 [30-39], 393 [40-49], 487 [50-59], 390 [60-69], 212 [70-79] and 39 [80-89]. 

\begin{figure}[ht]
\vskip 0.2in
\begin{center}
\centerline{\includegraphics[width=\columnwidth]{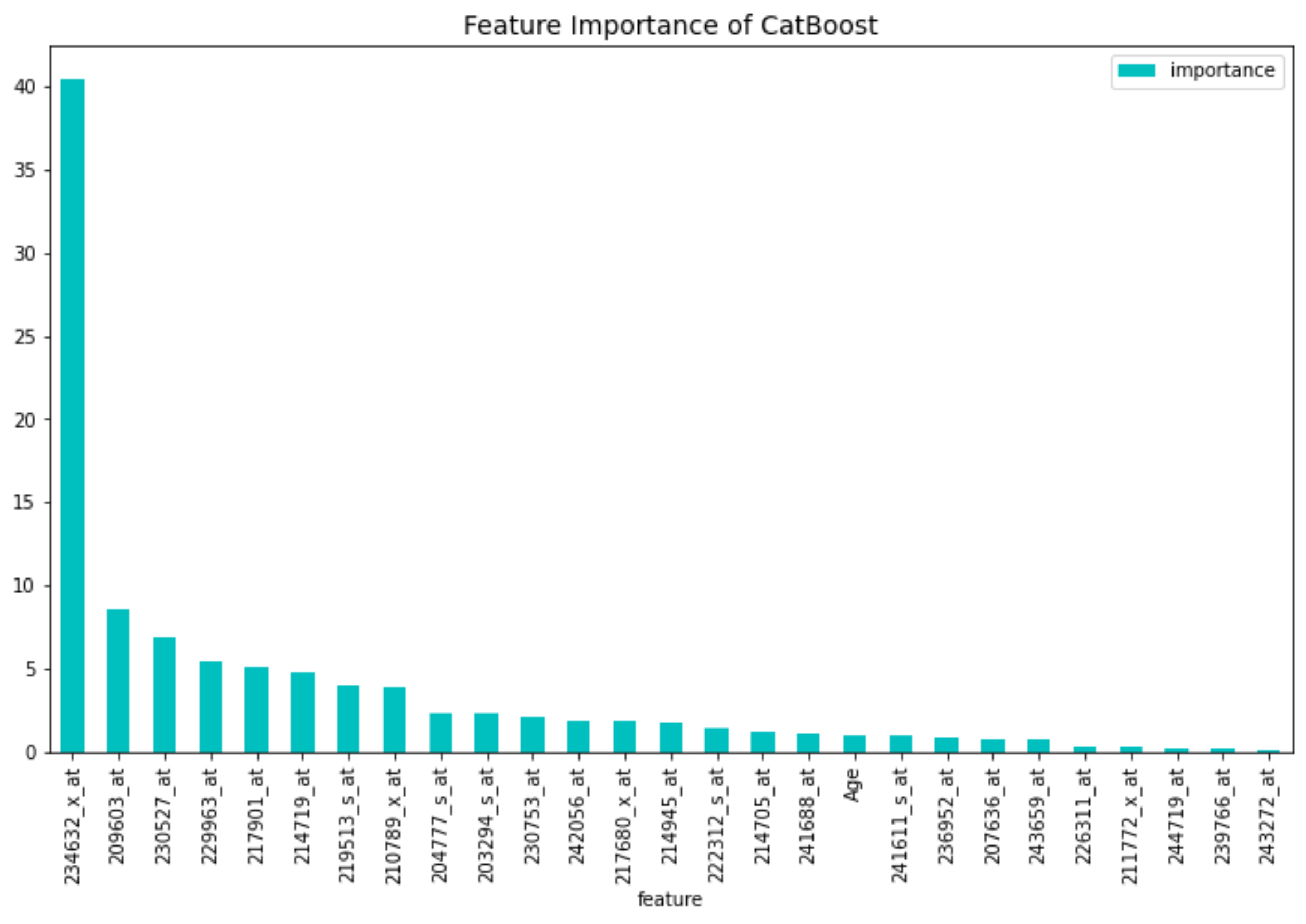}}
\caption{Features' importance of the predictability of CatBoost diagnosis model.}
\label{ficb}
\end{center}
\vskip -0.2in
\end{figure}

\begin{figure}[ht]
\vskip 0.2in
\begin{center}
\centerline{\includegraphics[width=\columnwidth]{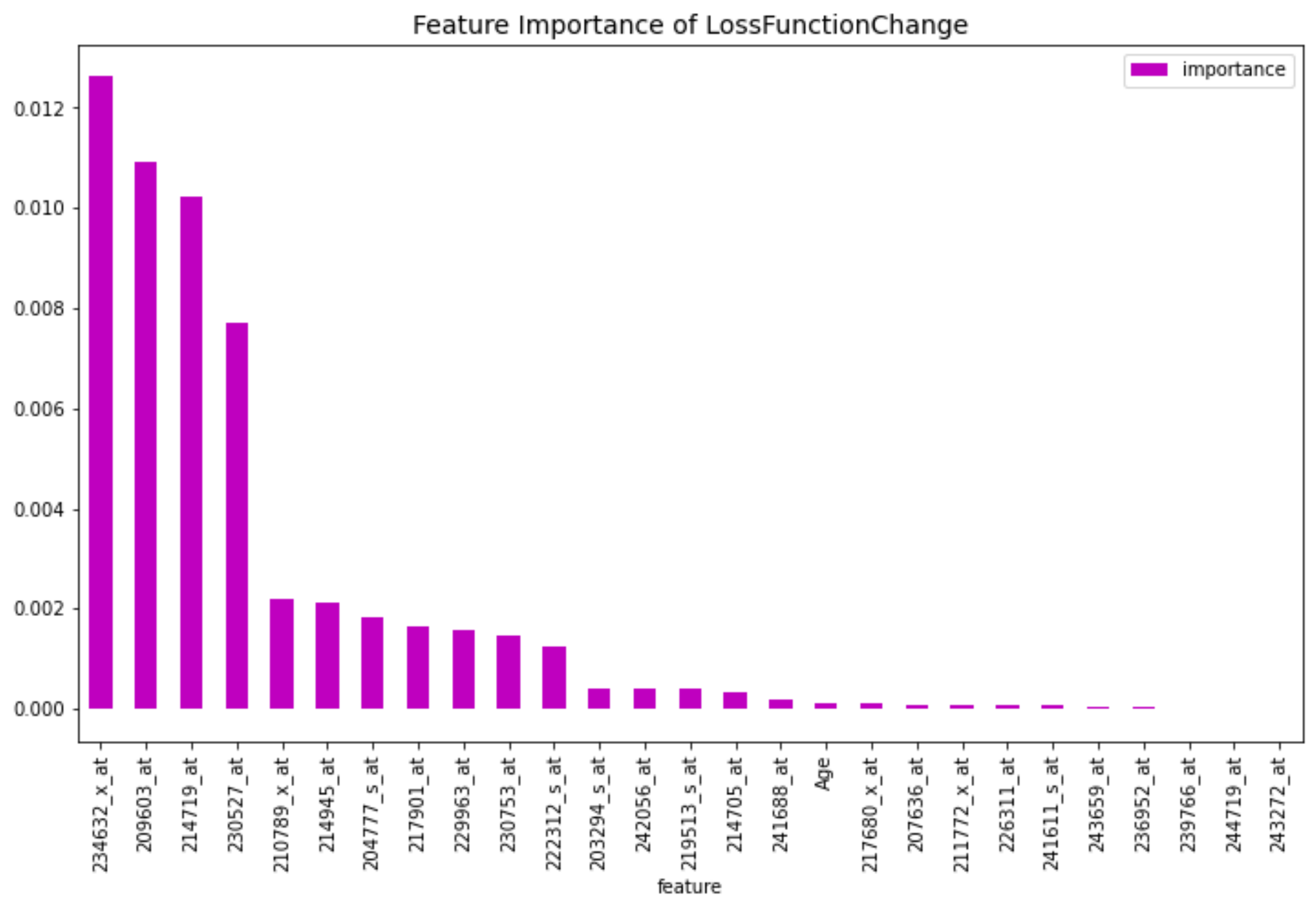}}
\caption{Features' importance of the CatBoost diagnosis model.}
\label{filf}
\end{center}
\vskip -0.2in
\end{figure}

We randomly split the final dataset in two sets: training and validation (Table \ref{trainset2177-table}, Table \ref{tage}). The training set consists of 1740 data instances (79.93\%) and the validation set of the rest 437 (20.07\%). Since the dataset is relatively small we use 10 fold cross validation in order to tune our model. In the Figure \ref{ficb} we observe the feature importance of the 27 features as regards the predictability of the CatBoost model using the 10CV, while in the Figure \ref{filf} we can see the feature importance of the loss change for each one of the 27 features. 

\section{Results}
At Table \ref{catboost_3-table} we see that our diagnosis model, CatBoost26, performs really well. The confusion matrix, Table \ref{cm-table}, shows the true-positives (down-right), the true-negatives (up-left), false-positives (down-left) and false-negatives (up-right). Here, a positive data instance is a data instance labelled as AML and negative as a healthy one.

\begin{table}[t]
\caption{Performance of the CatBoost diagnosis model, CatBoost26, of the 10CV on the 80\% training set and on the 20\% validation set of 2177 data instances on 26 probe sets and the age.}
\label{catboost_3-table}
\vskip 0.15in
\begin{center}
\begin{small}
\begin{sc}
\begin{tabular}{lcr}
\toprule
Metrics & Validation Set & 10CV \\
\midrule
Spec. & 1.0000 & 1.0000 \\
Sens. & 1.0000 & 0.9992\\
AUC   & 1.0000 & 0.9988\\
F1-score & 1.0000 & 0.9996\\
\bottomrule
\end{tabular}
\end{sc}
\end{small}
\end{center}
\vskip -0.1in
\end{table}

\begin{table}[t]
\caption{Confusion Matrix of the CatBoost diagnosis model's performance on the training set.}
\label{cm-table}
\vskip 0.15in
\begin{center}
\begin{small}
\begin{sc}
\begin{tabular}{lr}
\toprule
437 & 1 \\
0 & 1302\\
\bottomrule
\end{tabular}
\end{sc}
\end{small}
\end{center}
\vskip -0.1in
\end{table}

The mean area under the curve (AUC) from the 10CV is 0.9988 with standard deviation 0.0023 and 95\% confidence interval: $[0.9994, 1.000]$. The mean accuracy is 0.9994 with standard deviation 0.0011.

From Figures \ref{ficb} and \ref{filf} 
we observe that the probe set: 234632\_x\_at, which is ``cDNA'', is the most important probe set as regards both, the predictability of the CatBoost and the loss function change.

Our CatBoost34 model is transcriptomic platform agnostic \cite{Warnat20} since the label if a data instance comes either from Affymetrix Human Genome U133 Plus 2.0 microarray or the Affymetrix Human Genome U133A microarray or the Affymetrix Human Genome U133B microarray, has not been used as feature. This helps in the robustness and universality of our model's application in the diagnosis of AML. As regards the diagnosis model CatBoost26, all the data instances comes from the Affymetrix Human Genome U133 Plus 2.0.

From Figure \ref{ficb} we observe that the first 8 probe sets have the highest impact on the predictability of CatBoost26, including 6 named genes \{GATA3, BEX5, DSG2, SLC46A3, SH2D3A, CEACAM3\}, 1 uncharacterized gene \{LOC101926907\} and 1 cDNA probe set. The first probe set has remarkably high feature importance compared to the others, more than 4 times higher. To the best of our knowledge these genes have not been correlated to AML yet. The gene GATA3 has been correlated to acute lymphoblastic leukemia \cite{Hou17} and other types of cancer as well {breast cancer \cite{Mehra05}, bladder cancer \cite{Li14}}; the gene DSG2 is implicated in various kinds of cancer including cervical cancer \cite{Qin20}, epithelial-derived carcinomas \cite{Brennan09}, pancreatic cancer \cite{Hutz17}, breast cancer \cite{Davies97}, colon cancer \cite{Yang21}, lung cancer \cite{Cai17}, \cite{Saaber15}, gastric cancer \cite{Yashiro06}, \cite{Biedermann05}, ovarian cancer \cite{Kim20}, laryngeal cancer \cite{Cury20} and liver cancer \cite{Han18}. In addition, SLC46A3 is correlated to liver cancer \cite{Zhao19} and BEX5, SH2D3A, CEACAM3 have not been correlated to any type of cancer yet.

In Figure \ref{filf} we observe that the first 11 probe sets have the highest importance of loss function change of CatBoost26, including 10 named genes \{GATA3, BEX5, DSG2, SLC46A3, FAM153A, FAM153B, FAM153C, PATL2, CEACAM3, MAL\}, 3 uncharacterized genes \{LOC101926907, LOC100507387, LOC105377751\}, 1 expressed sequence tag and 1 cDNA probe set. The 
234632\_x\_at probe set has at least 4 times higher feature importance than the 210789\_x\_at, while 230527\_at is approximately 3 times more important feature than 210789\_x\_at. Moreover, the gene MAL has been correlated to gastric cancer \cite{Buffart08}, breast cancer \cite{Horne09}, ovarian cancer \cite{Lee10} and colorectal cancer \cite{Kalmar15}. The genes \{PATL2, FAM153A, FAM153B, FAM153C\} have not been correlated to any type of cancer yet.

\section{Related Work}
The first machine learning approach on a subset of the dataset of 3374 individuals with the 44754 probe sets, has been done in \cite{Roushangar19}. Statistical methods have been used in order to reduce the dimensionality of the dataset, which dropped down to 984 probe sets. Here we trained the k-NN machine learning model of \cite{Roushangar19} on the same 80\% train set as we did with our dimensionality reduction CatBoost model, using 10CV. The results, Table \ref{knn-table}, shows that the dimensionality reduction CatBoost model as well as CatBoost34, outperforms k-NN (Tables \ref{catboost_1-table}, \ref{catboost_2-table}).

\begin{table}[t]
\caption{Performance of the k-NN model of \cite{Roushangar19} of the 10CV on 80\% training set and on the 20\% validation set of 3374 data instances and 984 probe sets.}
\label{knn-table}
\vskip 0.15in
\begin{center}
\begin{small}
\begin{sc}
\begin{tabular}{lcr}
\toprule
Metrics & Validation Set & 10CV \\
\midrule
Spec. & 0.9716 & 0.9546 \\
Sens. & 0.9925 & 0.9920\\
AUC   & 0.9821 & 0.9788\\
F1-score & 0.9925 & 0.9899\\
\bottomrule
\end{tabular}
\end{sc}
\end{small}
\end{center}
\vskip -0.1in
\end{table}

Using similar to our work data of Affymetrix Human Genome U133A microarray, Affymetrix Human Genome U133 2.0 microarray and Illumina RNA-seq, different machine learning models and statistical learning techniques have been used (k-NN, LASSO, linear discriminant analysis, random forest, linear SVM, polynomial SVM, radial SVM, sigmoid SVM) in \cite{Warnat20} in order to predict if an individual has AML or is healthy. The best results regarding the accuracy are the following: 97.6\%, 98.0\% and 99.1\%. These results have been achieved by training and validating the LASSO algorithm on each of the Affymetrix Human Genome U133A microarray, Affymetrix Human Genome U133 2.0 microarray and Illumina RNA-seq datasets accordingly. The first dataset consisted of 2500 data instances from which 1049 (41.96\%) were labelled as AML and 1451 (58.04\%) as healthy. The second dataset consisted of 8348 data instances from which 2588 (31.00\%) were labelled as AML and 5760 (69\%) as healthy. Finally, the third dataset consisted of 1181 data instances from which 508 (43.01\%) were labelled as AML and 673 (56.99\%) as healthy. 

The last work related directly to ours is \cite{Nazari20} in which using microarrays a deep neural network (DNN) has been trained to classify AML from healthy individuals. The corresponding dataset consisted of only 26 data instances. DNN's accuracy score was 96.67\%.

All methods above use datasets from gene expression profiling (GEP) to diagnose AML. Another approach on different type of data like histopathology slides, using machine techniques has a  been tried out but the performance of the corresponding model, as regards accuracy, is around 95\% \cite{Kazemi16}.  

Using invariant cluster genomic signatures a machine learning approach has been developed in \cite{Awada21} for the classification of primary and secondary AML reaching an accuracy score of 97\%. 

Our method can be used in the classification of diffuse large B-cell lymphoma patients \cite{Zhao16} and also in sub-classification of leukaemia \cite{Castillo19} since GEP has been used as dataset in both cases.

\section{Conclusion}
We develop a model which using CatBoost and gene expression profiling data from Affymetrix Human Genome U133A, Affymetrix Human Genome U133B and Affymetrix Human Genome U133 2.0 is able to diagnose with the highest performance in literature if an individual has acute myeloid leukaemia or is healthy. We use CatBoost not only as a predictor to our problem, but also as a dimensionality reduction technique. In our approach both machine learning models, CatBoost34 and CatBoost26, outperform other machine learning approaches which use a variety of different classifiers and similar or different datasets.

On the clinical side, for the very first time in the literature we show that it is possible using probe sets, which have not been correlated yet to AML, and the state of the art machine learning gradient boosted tree algorithm CatBoost, to claim that we are able to diagnose AML. It would be of great importance to further investigate the role of these 26 probe sets, not only as regards the AML, but also other types of cancer. Machine learning can provide to us different insights from conventional approaches. As regards the explainability part, we hope the scientific community will use the importance of the probe sets shown in Figures \ref{ficb} and \ref{filf} in order to explain further their behavior in AML. In addition, from the 26 probe sets some of them have not been yet related to known genes. 

Acute myeloid leukaemia can appear suddenly to anyone. The importance of a screening tool where its sensitivity and specificity is close to 1.00, where the sample source is peripheral blood and the cost is low, it would have a tremendous impact to humanity. We hope our approach will inspire others to use machine learning in order to solve cancer problems. 

\section*{Acknowledgements}
We thank Michael Filippakis of University of Piraeus for his feedback and valuable discussions.

\onecolumn
\begin{table}[t]
\small
\caption{Number of samples and sample source, (peripheral blood (PB), bone marrow (BM)) of the train and validation set of 2177 individuals.}
\label{trainset2177-table}
\vskip 0.15in
\begin{center}
\begin{small}
\begin{sc}
\resizebox{\textwidth}{!}{%
\begin{tabular}{lcccccr}
\toprule
Index & Train \#Sam. \& Sam. Source & \% & Val. \#Sam. \& Sam. Source & \% \\
\midrule
0 & 6 BM & 75.00\% & 2 BM & 25.00\% \\ 
1 & 245 BM & 81.67\% & 55 BM & 18.33\% \\ 
2 & 22 PB & 84.62\% & 4 PB & 15.38\% \\
3 & 56 (52 BM \& 4 PB) & 71.80\% & 22 (21 BM \& 1 PB) & 28.20\% \\ 
4 & 412 (379 BM \& 33 PB) & 78.48\% & 113 (103 BM \& 10 PB) & 21.52\% \\ 
5 & 14 BM & 87.50\% & 2 BM & 12.50\% \\
6 & 194 (177 BM \& 17 PB) & 77.29\% & 57 (54 BM \& 3 PB) & 22.71\% \\ 
7 & 6 PB & 75.00\% & 2 PB & 25.00\% \\
8 & 18 PB & 81.82\% & 4 PB & 18.18\% \\ 
9 & 11 PB & 78.57\% & 3 PB & 21.43\% \\ 
10 & 13 PB & 76.47\% & 4 PB & 23.53\% \\ 
11 & 20 PB & 80.00\% & 5 PB & 20.00\% \\ 
12 & 50 PB & 79.37\% & 13 PB & 20.63\% \\ 
13 & 22 (12 BM \& 10 PB) & 64.71\% & 12 (9 BM \& 3 PB) & 35.29\% \\ 
14 & 11 PB & 91.67\% & 1 PB & 8.33\% \\
15 & 1 PB & 50.00\% & 1 PB & 50.00\% \\
16 & 11 (9 BM \& 2 PB) & 91.67\% & 1 BM & 8.33\% \\ 
17 & 28 PB & 80.00\% & 7 PB & 20.00\% \\
18 & 120 BM & 85.71\% & 20 BM & 14.29\% \\ 
19 & 37 PB & 80.43\% & 9 PB & 19.57\% \\
20 & 12 (10 BM \& 2 PB) & 92.30\% & 1 BM & 7.70\% \\
21 & 19 PB & 79.17\% & 5 PB & 20.83\% \\
22 & 9 (6 BM \& 3 PB) & 75.00\% & 3 (2 BM \& 1 PB) & 25.00\% \\ 
23 & 148 BM & 80.87\% & 35 BM & 19.13\% \\
24 & 12 PB & 100.00\% & - & 00.00\% \\
25 & 3 PB & 100.00\% & - & 00.00\% \\ 
26 & 42 (23 BM \& 19 PB) & 93.33\% & 3 (2 BM \& 1 PB) & 6.67\%\\ 
27 & 26 PB & 86.67\% &  4 PB & 13.33\% \\
28 & 25 PB & 71.43\% & 10 PB & 28.57\% \\
29 & 49 PB & 76.56\% & 15 PB & 23.44\% \\
30 & 99 PB & 81.82\% & 22 PB & 18.18\% \\ 
31 & - & 00.00\% & 1 PB & 100.00\% \\
\bottomrule
\end{tabular}
}
\end{sc}
\end{small}
\end{center}
\vskip -0.1in
\end{table}
\twocolumn

\onecolumn
\begin{table}[t]
\small
\caption{Number of patients per age group of the train and validation set. As regards the train set the mean of age is 48.98 with standard deviation 17.06 and as regards the validation set the mean of age is 48.46 and the standard deviation is 16.79.}
\label{tage}
\vskip 0.15in
\begin{center}
\begin{small}
\begin{sc}
\resizebox{\textwidth}{!}{%
\begin{tabular}{ m{5cm} m{2cm} m{5cm} m{2cm} }
\toprule
Train set & & Validation set&\\
Age group: \# Number of patients & \% & Age group: \# Number of patients & \%\\
\midrule
0 to 19: 75 & 4.31\% & 0 to 19: 24 & 5.5\% \\
20 to 29: 180 & 10.34\% & 20 to 29: 37 & 8.49\%\\
30 to 39: 272 & 15.62\% & 30 to 39: 68 & 15.6\%\\
40 to 49: 313 & 17.98\% & 40 to 49: 80 & 18.35\%\\
50 to 59: 378 & 21.71\% & 50 to 59: 109 & 25\%\\
60 to 69: 319 & 18.32\% & 60 to 69: 71 & 16.28\%\\
70 to 79: 171 & 9.82\% & 70 to 79: 41 & 9.4\%\\
80 to 100: 33 & 1.9\% & 80 to 100: 6 & 1.38\%\\
\bottomrule
\end{tabular}
}
\end{sc}
\end{small}
\end{center}
\vskip -0.1in
\end{table}
\twocolumn

\onecolumn
\begin{table}[t]
\small
\caption{References, GEO Accesion, Health Status, Origin of Study, AML subtypes of the study and Overall Survival of the train and validation set of 2177 individuals. The index corresponds to the index of Table \ref{trainset2177-table}. Some references which have note been published yet are notated with NYP.}
\label{info2177-table}
\vskip 0.15in
\begin{center}
\begin{small}
\begin{sc}
\resizebox{\textwidth}{!}{%
\begin{tabular}{m{3em} m{8em} m{3cm} m{4cm} m{6cm} m{3cm} m{2cm} }
\toprule
Index & Reference & GEO Acc. & AML/Healthy & City, Country, Org. & AML subtypes & OS \\
\midrule
0 & \cite{Zatkova09} & GSE10258 & AML & Vienna, Austria, Medical University of Vienna & M1, M5 & n/a \\

1 & \cite{Tomasson08}, \cite{Walter09} & GSE10358 & AML & St Louis, USA, Washington University School of Medicine & M0, M1,    & n/a \\
&  &  &  &  & M2, M3, & \\
&  &  &  &  & M4, M5, & \\
&  &  &  &  & M6, M7 & \\

2 & \cite{Warren09} & GSE11375 & Healthy & Boston, USA, Massachusetts General Hospital & n/a & n/a  \\

3 & \cite{Metzeler08}, \cite{Wang21} & GSE12417 & AML &  Munich, Germany, University of Munich & M0, M1,   & Mean: 614.76,  \\
&  &  &  &  & M2, M4, & Std: 503.59 \\
&  &  &  &  & M5, M6 & \\

4 & \cite{Wouters09}, \cite{Taskesen11}, \cite{Taskesen15}  & GSE14468 & AML & Houston, USA, MD Anderson Cancer Center & M0, M1,   & n/a\\
  &  &  &  &  & M2, M3, & \\
  &  &  &  &  &  M4, M4 eos, & \\
  &  &  &  &  & M5, M6 & \\
5 & \cite{Figueroa09} & GSE14479 & AML & Rotterdam, Netherlands, Erasmus University Medical Center & n/a & n/a \\

6 & \cite{Klein09}  & GSE15434 & AML & New York, USA, Columbia University Medical Center & n/a & n/a \\

7 & Wu 2012 (NYP) & GSE15932 & Healthy & Hangzhou, China, Second Affiliated Hospital, School of Medicine, Zhejiang University & n/a & n/a \\

8 & \cite{Karlovich09} & GSE16028 & Healthy & Basel, Switzerland, F.Hoffmannn/La Roche AG & n/a & n/a \\

9 & Krug 2011 (NYP) & GSE17114 & Healthy & Lisbon, Portugal, Instituto de Medicina Molecular & n/a & n/a \\

10 & \cite{Kong12}  & GSE18123 & Healthy & Boston, USA, Boston Children's Hospital & n/a & n/a \\

11 & \cite{Sharma09} & GSE18781 & Healthy & Portland, USA, Oregon Health \& Science University & n/a & n/a \\

12 & \cite{Zhou10} & GSE19743 & Healthy & Palo Alto, USA, Stanford Genome Technology Center & n/a & n/a \\

13 & \cite{Li11} & GSE23025 & AML & Duarte, USA, City of Hope Beckman Research Institute & n/a & n/a \\

14 & \cite{Rosell11} & GSE25414 & Healthy & Barcelona, Spain, Institut de Recerca Hospital Vall d'Hebron  & n/a & n/a \\

15 & \cite{Schmidt06} & GSE2842 & Healthy & Bolzano, Italy, EURAC & n/a & n/a \\

16 & \cite{Luck11} & GSE29883 & AML & Berlin, Germany, Charit\'{e} & t(8;21),  t(16;16) & n/a \\

17 & \cite{Xiao11} & GSE36809 & Healthy & Boston, USA, Massachusetts General Hospital & n/a & n/a \\

18 & \cite{Li13}, \cite{Herold14}, \cite{Kuett15}, \cite{Herold18} & GSE37642 & AML & Munich, Germany, University Hospital Grosshadern, Ludwign/Maximiliansn/University (LMU) & M0, M1,    & Mean: 962.32,  \\
&  &  &  &  & M2, M3, & Std: 1106.70 \\
&  &  &  &  & M4, M5, & \\
&  &  &  &  & M6, M7 & \\

19 & \cite{Lauwerys13}, \cite{Ducreux16} & GSE39088 & Healthy & Brussels, Belgium, Universit\'{e} catholique de Louvain & n/a & n/a \\

20 & Bullinger 2014 (NYP) & GSE39363 & AML & Berlin, Germany, Charit\'{e} & 
t(3;3) & n/a \\

21 & \cite{Clelland13}  & GSE46449 & Healthy & New York, USA, Columbia University Medical Center & n/a & n/a \\

22 & \cite{Opel15}, \cite{Lueck16} & GSE46819 & AML & Berlin, Germany, Charit\'{e} & t(16;16) & n/a \\

23 & Leong 2015 (NYP) & GSE68833 & AML & Rockville, USA, NCI & M0, M1, & n/a \\
&  &  &  &  & M2, M3, & \\
&  &  &  &  & M4, M5, & \\
&  &  &  &  & M6, M7 & \\

24 & \cite{Cao16} & GSE69565 & AML & Singapore, Singapore, Cancer Science Institute of Singapore  & n/a & n/a \\

25 & Meng 2015 (NYP) & GSE71226 & Healthy & Changchun, China, the Department of Cardiology, China-Japan Union Hospital, Jilin University  & n/a & n/a \\

26 & Bohl 2016 (NYP) & GSE84334 & AML & Ulm, Germany, University Hospital of Ulm & n/a & n/a \\
27 & \cite{Tasaki17} & GSE84844 & Healthy & Fujisawa, Japan, Takeda Pharmaceutical Company Limited  & n/a & n/a \\
28 & \cite{Tasaki18} & GSE93272 & Healthy & Fujisawa, Japan, Takeda Pharmaceutical Company Limited  & n/a & n/a \\
29 & \cite{Leday18} & GSE98793 & Healthy & Cambridge, United Kingdom, University of Cambridge  & n/a & n/a \\

30 & \cite{Shamir17} & GSE99039 & Healthy & Tel Aviv, Israel, Tel Aviv University  & n/a & n/a \\

31 & Green 2009 (NYP) & GSE14845 & Healthy & Southport, Australia, Griffith Insitute for Health \& Medical Research  & n/a & n/a \\

\bottomrule
\end{tabular}
}
\end{sc}
\end{small}
\end{center}
\vskip -0.1in
\end{table}
\twocolumn

\onecolumn
\begin{table}[t]
\small
\caption{The 26 probe sets ranked by their feature importance regarding the predictability of CatBoost. Information such as, probe set's ID, corresponding gene symbols or NCBI accession numbers, blood malignancies and/or other types of cancer they are associated with, as well as, general annotations about the probe sets and the role of the gene products are presented here.}
\label{t26-probesets}
\vskip 0.15in
\begin{center}
\begin{small}
\begin{sc}
\resizebox{\textwidth}{!}{%
\begin{tabular}{m{8em} m{8em} m{4cm} m{4cm} m{7cm} }

\toprule
Probe set ID & Gene Symbol/NCBI Accesion Number & Blood Malignancies & Other types of cancer & General \\
\midrule
234632\_x\_at & AK026267 & n/a & n/a & cDNA: FLJ22614 fis, clone HSI05089 \cite{Wheeler05} \\
209603\_at & GATA3 & Acute Lymphoblastic Leukemia (ALL)\cite{Hou17} & Breast Cancer \cite{Mehra05}, Bladder Cancer \cite{Li14} & This gene encodes a protein, which plays a role as regulator of T-cell development \cite{Wheeler05} \\
230527\_at & LOC101926907 & n/a & n/a & Uncharacterized Gene \cite{Wheeler05} \\
229963\_at & BEX5 & n/a & n/a & The protein encoded by this gene plays a role in neuronal development \cite{Kazi15} \\
217901\_at & DSG2 & n/a & Cervical Cancer \cite{Qin20},
Epithelial-derived Carcinomas \cite{Brennan09},
Pancreatic Cancer \cite{Hutz17}, Breast Cancer \cite{Davies97}, Colon Cancer \cite{Yang21}, Lung Cancer \cite{Cai17}, \cite{Saaber15}, Gastric Cancer \cite{Yashiro06}, \cite{Biedermann05}, Ovarian Cancer \cite{Kim20}, Laryngeal Cancer \cite{Cury20}, Liver Cancer \cite{Han18} & This gene encodes a calcium-binding transmembrane glycoprotein component of desmosomes, which plays a role in cell-cell junctions between epithelial, myocardial, and other types of cells \cite{Wheeler05} \\
214719\_at & SLC46A3 & n/a & Liver Cancer \cite{Zhao19} & This gene encodes a protein, which is involved in transportation of small molecules across membranes \cite{Wheeler05} \\
219513\_s\_at & SH2D3A & n/a & n/a & This gene encodes a protein, which may play a role in JNK activation \cite{Bateman21} \\
210789\_x\_at & CEACAM3 & n/a & n/a & The protein encoded by this gene it is thought to play an important role in controlling human-specific pathogens \cite{Wheeler05} \\
204777\_s\_at & MAL & n/a & Gastric Cancer \cite{Buffart08}, Breast Cancer \cite{Horne09}, Ovarian Cancer \cite{Lee10}, Colorectal Cancer \cite{Kalmar15} & This gene encodes a protein, which plays a central role in the formation, stabilization and maintenance of glycosphingolipid-enriched membrane microdomains \cite{Wheeler05} \\
203294\_s\_at & LMAN1 & n/a & n/a & This gene encodes a protein, which is involved in glycoprotein transportation \cite{Wheeler05} \\
230753\_at & PATL2 & n/a & n/a & This gene encodes an RNA-binding protein, which plays a role as translational repressor in regulation of maternal mRNAs during oocyte maturation \cite{Cao21} \\
242056\_at & TRIM45 & n/a & Lung Cancer \cite{Peng19}, Glioma \cite{Zhang17} & The encoded protein acts as a transcriptional repressor of the mitogen-activated protein kinase pathway \cite{Wheeler05} \\
217680\_x\_at & RPL10 & T-cell Acute Lymphoblastic Leukemia (T-ALL) \cite{Raiser14}, \cite{De13} & Ovarian Cancer \cite{Shi18}, Pancreatic Cancer \cite{Yang18} & The encoded protein is a component of the 60S ribosome subunit \cite{Wheeler05} \\
214945\_at & FAM153A \& FAM153B \& FAM153C \& LOC100507387 \& LOC105377751 & n/a & n/a & Unknown function/Uncharacterized gene \cite{Wheeler05} \\
222312\_s\_at & AW969803 & n/a & n/a & Expressed sequence tag \cite{Wheeler05} \\
214705\_at & PATJ & n/a & n/a & This gene encodes a protein, which is located in tight junctions and in the apical membrane of epithelial cells \cite{Wheeler05} \\
241688\_at & AA677700 & n/a & n/a & Expressed sequence tag \cite{Wheeler05} \\
241611\_s\_at & FNDC3A & Multiple Myeloma \cite{Manfrini20} & n/a & The protein encoded by this gene plays a role in spermatid-Sertoli adhesion during spermatogenesis \cite{Obholz06} \\
236952\_at & AI309861 & n/a & n/a & Expressed sequence tag \cite{Wheeler05} \\
207636\_at & SERPINI2 & Chronic Lymphocytic Leukemia (CLL) \cite{Farfsing09} & pancreatic cancer \cite{Ozaki98} & The encoded protein is involved in the regulation of a variety of physiological processes, including coagulation, fibrinolysis, development, malignancy, and inflammation \cite{Wheeler05} \\
243659\_at & N63876 & n/a & n/a & Expressed sequence tag \cite{Wheeler05} \\
226311\_at & ADAMTS2 & Mixed Phenotype Acute Leukemias (MPAL) \cite{Tota19} & Gastric Cancer \cite{Jiang19}, Kidney Cancer \cite{Roemer04} & This gene encodes an extracellular metalloproteinase,
which plays a significant role in the excision of the N-propeptides of procollagens I-III and type V \cite{Wheeler05} \\
211772\_x\_at & CHRNA3 & T-cell Acute Lymphoblastic Leukemia (T-ALL) \cite{Laukkanen15} & Lung Cancer \cite{Wassenaar11} & The protein encoded by this gene is a ligand-gated ion channel,
which plays a role in neurotransmission \cite{Wheeler05} \\
244719\_at & AA766704 & n/a & n/a & Expressed sequence tag \cite{Wheeler05} \\
239766\_at & BF507518 & n/a & n/a & Expressed sequence tag \cite{Wheeler05} \\
243272\_at & LOC101593348 & n/a & n/a & Uncharacterized gene \cite{Wheeler05} \\
\bottomrule
\end{tabular}
}
\end{sc}
\end{small}
\end{center}
\vskip -0.1in
\end{table}
\twocolumn



\end{document}